\documentclass{article}
\usepackage{ijcai24}

\usepackage{times}
\usepackage{soul}
\usepackage{url}
\usepackage[hidelinks]{hyperref}
\usepackage[utf8]{inputenc}
\usepackage[small]{caption}
\usepackage{graphicx}
\usepackage{amsmath}
\usepackage{amsthm}
\usepackage{booktabs}
\usepackage[switch]{lineno}

\usepackage{amssymb}
\usepackage[linesnumbered,ruled]{algorithm2e}
\usepackage{algpseudocode}
\usepackage{tabularx}
\usepackage{diagbox} 
\usepackage{longtable}
\usepackage{multirow}
\usepackage{makecell}
\usepackage{subcaption}
\usepackage{todonotes}
\usepackage{enumitem}

\title{Federated Unlearning for Human Activity Recognition}

\author{
Kongyang Chen$^1$ \and
Dongping zhang$^1$\and
Yaping Chai$^{1}$\and
Weibin Zhang$^{2}$\and \\
Shaowei Wang$^{1,*}$\and
Jiaxing Shen$^{3,*}$\\
\affiliations
$^1$Guangzhou University  $^2$Xidian University  $^3$Lingnan University\\
\emails
\{kychen, wangsw\}@gzhu.edu.cn,
jiaxingshen@ln.edu.hk
}

\begin{document}
\maketitle

\begin{abstract}
The rapid evolution of Internet of Things (IoT) technology has spurred the widespread adoption of Human Activity Recognition (HAR) in various daily life domains. Federated Learning (FL) is frequently utilized to build a global HAR model by aggregating user contributions without transmitting raw individual data. Despite substantial progress in user privacy protection with FL, challenges persist. Regulations like the General Data Protection Regulation (GDPR) empower users to request data removal, raising a new query in FL: How can a HAR client request data removal without compromising other clients' privacy? In response, we propose a lightweight machine unlearning method for refining the FL HAR model by selectively removing a portion of a client's training data. Our method employs a third-party dataset unrelated to model training. Using KL divergence as a loss function for fine-tuning, we aim to align the predicted probability distribution on forgotten data with the third-party dataset. Additionally, we introduce a membership inference evaluation method to assess unlearning effectiveness. Experimental results across diverse datasets show our method achieves unlearning accuracy comparable to \textit{retraining} methods, resulting in speedups ranging from hundreds to thousands.
\end{abstract}

\section{Introduction}
Human Activity Recognition (HAR) represents a significant area of research within the realms of Artificial Intelligence and signal processing. The primary aim of HAR is to discern and categorize human activities through data gathered from various sensors, including accelerometers and gyroscopes. HAR finds diverse applications in areas such as health monitoring, fall detection, and smart homes \cite{rosaline2023enhancing}, \cite{yadav2022arfdnet}, \cite{bianchi2019iot}. Traditional HAR methodologies typically depend on centralized training of global models using data sourced from different users or devices. However, this centralized approach engenders concerns related to privacy, security, and scalability \cite{zhou2014learning}. To mitigate these issues, Federated Learning (FL) \cite{konevcny2016federated}, \cite{mcmahan2017communication} has surfaced as a promising solution, facilitating local training on individual devices without necessitating the sharing of raw data.

Nonetheless, FL continues to grapple with security and privacy issues within the context of HAR. From a privacy viewpoint, HAR datasets, collected via sensors positioned on the human body, are inherently sensitive. From a security viewpoint, the evolving techniques of malicious attackers enable them to extract pertinent data from the globally trained model on participating clients \cite{fredrikson2015model}, \cite{salem2018ml}, \cite{shokri2017membership}, \cite{zhang2020secret}. Regardless of whether it is before or after model training, participating clients must address security and privacy concerns related to their data with utmost caution. To bolster the protection of public and personal data, numerous countries and organizations have implemented regulations such as the General Data Protection Regulation (GDPR) \cite{voigt2017eu} and the California Consumer Privacy Act (CCPA) \cite{pardau2018california}, thereby granting users the \textit{right to be forgotten}. Consequently, after training the global model using their data, FL must afford participating clients the right to erase their data from the model.

Prior to participating in model training, clients can preprocess the training data utilizing privacy-preserving techniques such as Differential Privacy (DP) \cite{dwork2006differential}. However, within the context of HAR under FL, the model's generalization ability is impeded due to the heterogeneity of each client's data, which originates from individuals with varying factors such as environment, age, and habits. The incorporation of noise in DP further diminishes the accuracy of the training model due to these disparities. Post participation in model training, clients can employ Machine Unlearning \cite{cao2015towards} technology to forget data that could potentially pose privacy risks from the model, a process referred to as Federated Unlearning (FU) in FL. FU enables clients that partake in global model training to remove their data from the model. Moreover, to mitigate data security and privacy concerns, clients may also request the deletion of their data from the model subsequent to participating in global model training. The most straightforward method to forget client data is to negate the contribution of this data from the model through retraining. Generally, FU technology encompasses three common unlearning scenarios: unlearning client training data subsets, unlearning categories, and unlearning clients themselves (e.g., when the client exits training or is attacked) \cite{yang2023survey}. Given that the server cannot directly access the data for the unlearning scenario of client training data subsets, the unlearning process must be executed locally on the client. Upon completion of the unlearning, the server updates all clients' models with the unlearned data model to achieve the unlearning effect. If unlearning is accomplished through retraining, the server needs to aggregate the model trained by the clients and deploy a new model to them, resulting in substantial communication overhead. Utilizing \textit{retrain} methods exclusively for unlearning a portion of client data is resource-inefficient. However, to the best of our knowledge, there is currently no established solution for the (Federated) unlearning process in HAR.

In this paper, we would like to address the clients' concerns regarding the privacy of their data subsequent to participating in global model training, and their potential request for the server (service provider) to erase their training data that could potentially pose privacy risks in the model. Consequently, we propose a method to fine-tune the global model in the FL HAR scenario to unlearn a portion of the client's training data. Our method, in contrast to the retrain method, conserves a significant amount of communication resources and does not necessitate the participation of other clients. To accomplish this, our method employs a third-party dataset $D_{t}^{i}$ that does not partake in the model training. We use KL divergence as a loss function to fine-tune the model, with the objective of aligning the predicted probability distribution on the forgotten data $D_{f}^{i}$ with that of $D_{t}^{i}$. Additionally, we introduce a membership inference evaluation method to validate the effectiveness of unlearning. Experimental results across diverse datasets demonstrate that our approach attains comparable accuracy in unlearning when compared to \textit{retraining} methods, resulting in a speedup ranging from $294\times$ to $6119\times$. The main contributions of our work can be summarized as follows:

\begin{itemize}
    \item We propose a lightweight unlearning method designed to fine-tune models for the purpose of discarding a subset of data within the Federated Learning Human Activity Recognition (FL HAR) scenario. We also introduce a method for membership inference evaluation to validate the efficacy of the unlearning process. Unlike the \textit{retraining} method, our approach significantly conserves computational resources.
    \item Our approach is applied to two HAR datasets and the MNIST dataset. Experimental results across these datasets show that our method achieves accuracy in unlearning comparable to \textit{retraining} methods, resulting in speedups ranging from hundreds to thousands.
\end{itemize}

The paper is organized as follows: Section \ref{sec:rw} provides an overview of recent works in the field. Section \ref{sec:pf} formulates the problem. Section \ref{sec:method} introduces the unlearning framework and presents our method. Section \ref{sec:evaluation} evaluates the effectiveness of our method. Finally, Section \ref{sec:conclusion} concludes this paper.

\section{Related work}
\label{sec:rw}
\textbf{HAR with Federated Learning}. Numerous studies have been conducted on HAR within the field of FL. To address the heterogeneity of user data in HAR scenarios, Tu et al. \cite{tu2021feddl} proposed a method known as FedDL, which discerns the similarity between user model weights and dynamically shares these weights, thereby accelerating convergence while preserving high accuracy. Li et al. \cite{li2021meta} discovered that heterogeneity in label and signal distribution significantly impacts the efficacy of FL for HAR. They incorporated meta-learning into FL and introduced Meta-HAR, which effectively enhances model personalization performance. To simultaneously address privacy preservation, label scarcity, real-time processing, and heterogeneous patterns of HAR, Yu et al. \cite{yu2021fedhar} proposed a semi-supervised personalized FL framework named FedHAR. Ouyang et al. \cite{ouyang2021clusterfl}, by identifying intrinsic similarities among the activity data of users, proposed ClusterFL, which performs clustering learning based on user similarities and introduced two mechanisms to augment accuracy and reduce communication overhead. Shen et al. \cite{shen2022federated} noted that individuals typically perform identical activities in varying ways in real-life scenarios. They proposed a federated multi-task attention framework named FedMAT, which addresses HAR challenges by extracting features from shared and personal-specific data. An FL framework needs to consider various aspects, including accuracy, robustness, fairness, and scalability. Therefore, Li et al. \cite{li2023hierarchical} proposed the FedCHAR framework, which emphasizes robustness and fairness, and further introduced the scalable and adaptive FedCHAR-DC. However, previous works, while considering security and privacy issues, focused on preprocessing prior to model training and did not address the security and privacy concerns that persist post model training.

\textbf{Machine Unlearning}. The concept of the \textit{right to be forgotten}, underscored in regulations such as GDPR, highlights the user's right to not only delete data from storage devices but also to earse it from learned models. While data removal from storage is a straightforward process, it becomes complex when it entails eradicating data from a trained model.
Cao et al. \cite{cao2015towards} initially coined the term \textit{Machine Unlearning} to depict the process of forgetting data from a learning system (model). They transformed the learning algorithm of the system (model) into a summation form, necessitating only minimal updates to forget training data samples.
Golatkar et al. \cite{golatkar2020eternal} defined the forgetting objective as $KL(M_t||M_u)$ symbolize the target model to forget (i.e., $M_t$) and the model post forgetting (i.e., $M_u$), respectively. The optimal forgetting, termed \textit{Exact Unlearning} by Xu et al. \cite{xu2023machine}, is achieved when $KL(M_t||M_u)=0$ and \textit{Approximate Unlearning} is realized when $KL(M_t||M_u)$ approaches 0.
Chen et al. \cite{chen2021machine} proposed a GAN-based unlearning algorithm capable of rapidly erasing forgotten data from the model.

\textbf{Federated Unlearning}. The procedure of unlearning data in a FL model is termed Federated Unlearning (FU). The unlearning challenge in FL slightly deviates from that in centralized learning models.
Yang et al. \cite{yang2023survey} classified the FU problem into three levels of unlearning: sample-level, class-level, and client-level. This involves forgetting a portion of client data, a category of data, and forgetting an entire client.
In previous works, the emphasis has been primarily on how to earse clients' contributions (in cases of client dropout or client attacks). \cite{wu2022federated}, \cite{yuan2023federated}, \cite{halimi2022federated}, \cite{zhang2023fedrecovery} used server-stored historical updates from clients to forget client contributions in the model, thereby achieving the forgetting of specific clients.
Su et al. \cite{su2023asynchronous} employed clustering to group participating clients, executing the unlearning process solely within the cluster where the client resides, thus minimizing the impact of unlearning on the model.
Wang et al. \cite{wang2023bfu} introduced the BFU (BFU-SS) method, which divides data unlearning and model accuracy preservation into two tasks, optimizing them concurrently during the unlearning process. The unlearning process is executed in tandem with FL, and the server remains oblivious of the unlearning process. 

As far as our knowledge extends, there is currently no established solution for the unlearning process in (Federated) HAR. We hope that our work will be a pioneering unlearning solution for HAR, contributing to enhanced privacy protection in this area.

\section{Problem formulation}
\label{sec:pf}
Federated Learning, as a decentralized machine learning approach, bears significant relevance to users in Human Activity Recognition (HAR) scenarios. HAR data, being a rich source of information about human activities, is inherently more sensitive and personal. It includes a wide range of human behaviors, from simple actions like walking or standing to complex activities like exercising or operating machinery. Given the intimate nature of this data, it carries a high risk of privacy invasion if mishandled or misused. Therefore, in the context of HAR, privacy protection should be prioritized. Implementing FL in such scenarios can provide a robust framework for learning from this sensitive data while ensuring that the privacy of individuals is respected and protected.

Assume that $n$ clients are participating in model training. Let $C=\{C_i\}_{i=0}^{n-1}$ denotes the set of these $n$ clients. The training data for client $C_i$ is represented by $D_{c}^{i}=\{(x,y)\}$, where $x\in\mathcal{X}\subset\mathbb{R}^d$ is the input data, and $y\subset\mathcal{Y}=0,1,\ldots,K$ is the corresponding label. The model trained by client $C_i$ using the dataset $D_{c}^{i}$ with the learning algorithm $\mathcal{A}_c(\cdot)$ is denoted as $M_{c}^{i}$. In this context, $\mathcal{A}_c(\cdot)$ signifies the learning algorithm employed by the client, and $M_{c}^{i}=\mathcal{A}_c(D_{c}^{i})$. The server model $M_s$, is aggregated using the $FedAvg$ algorithm \cite{mcmahan2017communication}, which amalgamates the models uploaded by the clients.
\begin{equation}
    M_s=FedAvg(M_{c}^{0},M_{c}^{1},\dots,M_{c}^{n-1}).
\end{equation}
During the initial training phase, the models of the clients are initialized to be equivalent to the server model, denoted as $M_{c}^{i}=M_s$, where $i=0,1,\dots,n-1$.

When client $C_i$ requests to unlearn $D_{f}^{i}$ from the global model $M_{s}$, where $D_{f}^{i} \cup D_{r}^{i} = D_{c}^{i}$, $D_{f}^{i} \cap D_{r}^{i} = \emptyset$, we define $U(.)$ as the unlearning algorithm. The function $U(.)$ is responsible for unlearning $D_{f}^{i}$ from $M_{s}$.
\begin{equation}
    M_u=U(M_s,D_{f}^{i}). \label{eq:mu1}
\end{equation}
$M_u$ is the model obtained after applying the unlearning algorithm $U(\cdot)$. The metric for evaluating the algorithm $U(\cdot)$ is that the model $M_u$ should not contain any information about $D_{f}^{i}$.

To remove a portion of data from the model, it is natural to consider using the \textit{retrain} method with the remaining data. For instance, if an unlearning request is sent to the server using $C_0$, the retraining method for unlearning necessitates that the client request the server to retrain the model. $C_0$ uses $D_{r}^0$ for training, while the other clients use the complete dataset $D_{c}^i$ to retrain the model.
\begin{equation}
    M_u=FedAvg(\mathcal{A}_c(D_{r}^{0}),\mathcal{A}_c(D_{c}^{1}),\ldots,\mathcal{A}_c(D_{c}^{n-1})).
    \label{eq:mu2}
\end{equation}

Nonetheless, within a federated environment, the \textit{retrain} method encounters more challenges compared to centralized learning approaches. If we assume that the clients involved in the training frequently issue unlearn requests, these clients would necessitate substantial retraining, leading to significant resource consumption. Moreover, for clients who do not need to unlearn data, their repeated involvement in unrelated training could potentially result in their reluctance to participate in the retraining process.

\section{Our method}
\label{sec:method}
In this paper, we propose an efficient and privacy-preserving method for unlearning FL models within a federated setting. Our approach leverages third-party data that is not involved in the model training process to fine-tune the global model and achieve data forgetting. This section begins with an overview of our motivation, followed by a comprehensive explanation of the design of our method. We then delve into the issue of selecting appropriate third-party data. Lastly, we introduce the technical methods employed to evaluate the effectiveness of our proposed approach.

\subsection{Federated unlearning framework}
\begin{figure}[!t]
    \centering
    \includegraphics[width=0.85\linewidth]{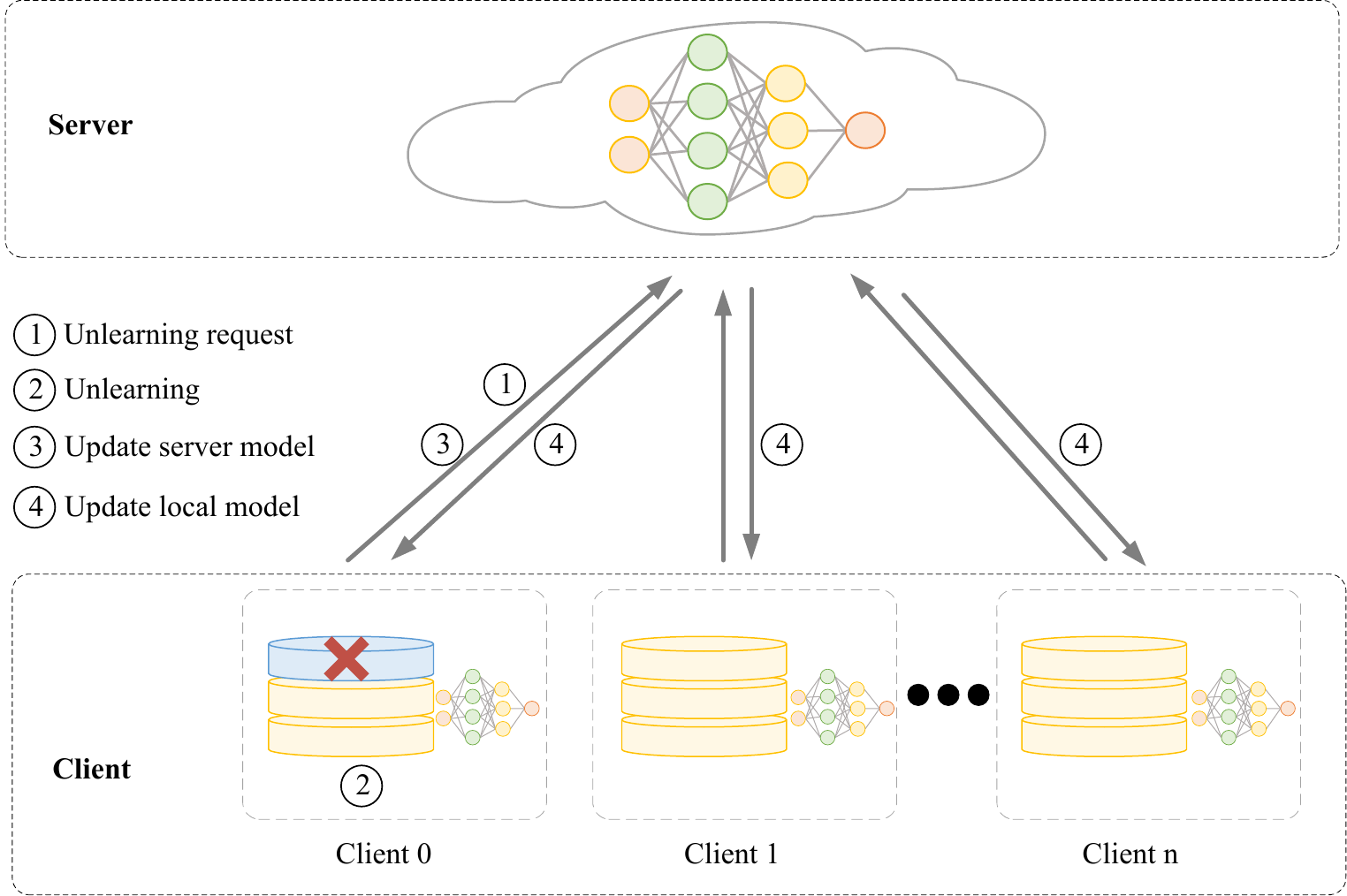}
    \caption{Federated unlearning framework for HAR.}
    \label{fig:fu}
\end{figure}

Machine Learning models have encountered the issue of memorizing training data during the training process \cite{shokri2017membership}, \cite{salem2018ml}, \cite{carlini2019secret}. In light of concerns regarding privacy and security during participation in global model training, clients may request to remove specific data from the model. The primary process of unlearning client data is illustrated in Figure \ref{fig:fu}, encompassing the following steps: 1) the client sends an unlearning request to the server; 2) upon receiving the request, the server updates the client's model, enabling the client to execute the unlearning operation; 3) after completing the unlearning process, the client transmits the post-unlearning model to the server, which subsequently updates its own model; and 4) the server then disseminates the post-unlearning model to all clients, ensuring that neither the server nor any client's model contains $D_{f}^{i}$. To provide enhanced privacy protection, the unlearning process for clients should be executed locally. Consequently, in a federated scenario, the post-unlearning client model needs to be uploaded to the server, which then updates the local models of all clients to achieve true unlearning success within the federated environment.

The concept of unlearning involves treating trained data as untrained data. Motivated by this insight, we propose the utilization of Kullback-Leibler divergence (KL divergence) \cite{kullback1997information} for achieving unlearning. The underlying principle of unlearning through KL divergence is to minimize the discrepancy between the predicted distribution of the unlearned data $D_{f}^{i}$ and the predicted distribution of third-party data $D_{t}^{i}$, which remains untrained. This objective can be formulated as a loss function and optimized using the stochastic gradient descent algorithm to fine-tune the model's prediction probabilities for $D_{f}^{i}$ and facilitate unlearning. Equation \ref{eq:kla} represents the optimization target when employing KL divergence as the loss function.
\begin{equation}
    \label{eq:kla}
    \arg\min_{M_{u}} KL(Pr(M_{u}(D_{f}^{i})) \| Pr(M_{u}(D_{t}^{i}))).
\end{equation}

Although Equation \ref{eq:kla} can achieve unlearning for $D_{f}^{i}$, it may result in a performance decline on $D_{r}^{i}$. The key to successful unlearning lies in maintaining the model's performance while achieving the desired unlearning effect. To preserve the overall performance, optimization of Equation \ref{eq:kla} needs to be performed. We propose incorporating the client's remaining data $D_{r}^{i}$ into the unlearning process and optimizing the model to minimize the loss on the remaining data $M_u$, thereby striking a balance in the model's performance after unlearning. The optimization objective is expressed as follows:
\begin{equation}
    \label{eq:loss}
    \arg\min_{M_{u}}\mathcal{L}(M_{u}(D_{r}^{i})).
\end{equation}

We introduce a hyperparameter $\lambda$ to incorporate both Equation \ref{eq:kla} and Equation \ref{eq:loss}. This hyperparameter $\lambda$ acts as a weight to strike a balance between the unlearning effect and the model's performance on the remaining dataset. When $\lambda=1$, it signifies a sole focus on the unlearning effect. The final optimization objective is formulated as follows:
\begin{equation}
    \label{eq:ua}
    \begin{split}
        \arg\min_{M_{u}} \lambda KL(Pr(M_{u}(D_{f}^{i}))\| Pr(M_{u}(D_{t}^{i}))) \\ +(1-\lambda) \mathcal{L}(M_{u}(D_{r}^{i})).
    \end{split}
\end{equation}

\subsection{Federated unlearning method}
\begin{figure}[!t]
    \centering
    \includegraphics[width=0.85\linewidth]{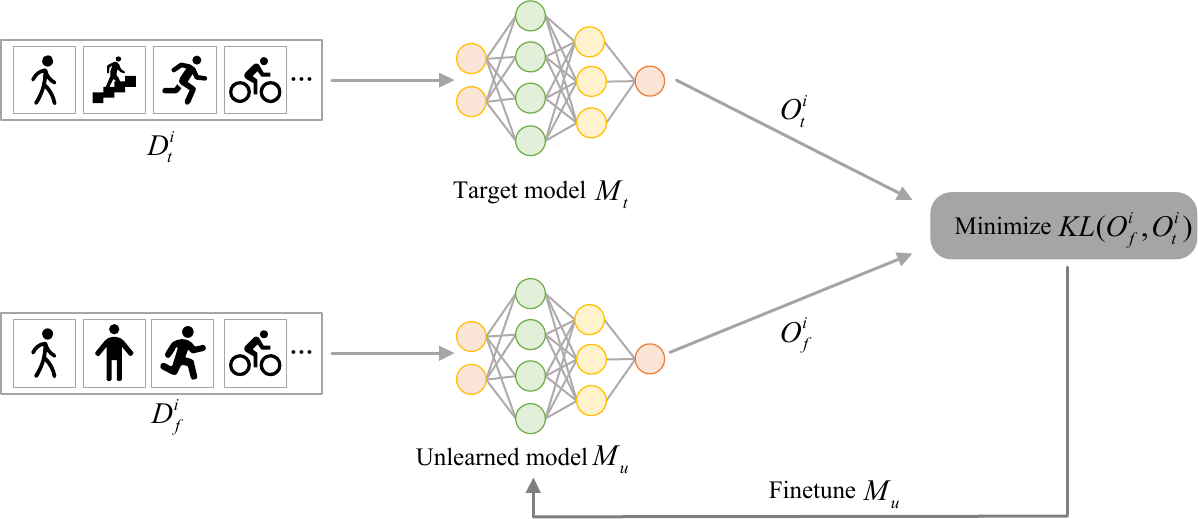}
    \caption{Federated unlearning method.}
    \label{fig:unlearning}
\end{figure}

The KL unlearning algorithm executed by the client is presented in Figure \ref{fig:unlearning}. Upon sending an unlearning request to the server, the client receives the latest global model $M_t$. Subsequently, the client initializes $M_u$ as $M_t$, representing the model after unlearning operations. The client then uses data $D_{t}^{i}$ and $D_{f}^{i}$ as inputs for models $M_t$ and $M_u$ respectively, generating outputs $O_{t}^{i}$ and $O_{f}^{i}$. These outputs are fed into the KL function, which fine-tunes model $M_u$ by minimizing the distribution difference between $O_{t}^{i}$ and $O_{f}^{i}$. The algorithm terminates when the distribution difference between $O_{t}^{i}$ and $O_{f}^{i}$ becomes sufficiently small. Algorithm \ref{alg:ula} provides the pseudo-code for the client's unlearning of partial data, excluding the steps involving the unlearning request sent to the server and the server's update of all client models.

\begin{algorithm}[!t]
    \caption{Federated unlearning algorithm}
    \label{alg:ula}
    \KwData{Global model $M_{t}$; $C_{i}$ unlearn data $D_{f}^{i}$; $C_{i}$ remaining data $D_{r}^{i}$; $C_{i}$ prepared third-party data $D_{t}^{i}$; Hyperparameters $\lambda$; Unlearning epochs $E$.}
    \KwResult{Unlearned model $M_u$.}
    \SetAlgoNlRelativeSize{}
    \SetNlSty{}{}{:}
    \textbf{}
    Initialize the unlearned model$M_u=M_{t}$.\\
    \For{$t\leftarrow 1$ \KwTo $E$}{
    Sample minibatch $\{x_f\}_{i=1}^m$ from $D_{f}^{i}$;\\
    Sample minibatch $\{x_t\}_{i=1}^m$ from $D_{t}^{i}$;\\
    Sample minibatch $\{x_r,y_r\}_{i=1}^m$ from $D_{r}^{i}$;\\
    Update $M_u$ by descending its stochastic gradient:
    \begin{equation*}
        \begin{split}
            \mathcal{K}=KL\left(\mathrm{Pr}(M_{u}(x_{f}))\parallel\mathrm{Pr}(M_{t}(x_{t}))\right);\\
            \arg\min_{M_{u}}\lambda \mathcal{K}+(1-\lambda)\mathcal{L}\left(M_{u}(x_{r}),y_{r}\right);
        \end{split}
    \end{equation*}
    }
    \Return{The unlearned model $M_u$.}
\end{algorithm}

\subsection{Membership inference evaluation}
\begin{figure*}[!t]
    \centering
    \includegraphics[width=0.85\linewidth]{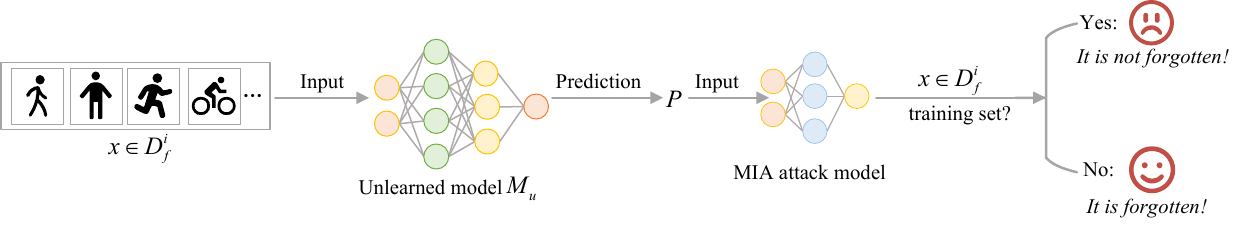}
    \caption{Membership inference evaluation method.}
    \label{fig:mia}
\end{figure*}

To verify the effectiveness of the unlearning process on dataset $D_{f}^{i}$, it is necessary to employ a technique known as Membership Inference Attack (MIA) \cite{shokri2017membership}. In the context of MIA, an independent third-party dataset $D_{out}^{i}$ is introduced. In this case, the target model $M_{t}$ to be unlearned is designated as the shadow model, as the goal is not to create an actual attack model. Prior to performing the unlearning operation on client $C_i$, the shadow model $M_{t}$ is trained using data $D_{in}^{i}$, and its output on $D_{in}^{i}$ and $D_{out}^{i}$ is denoted as $P_{in}^{i}$ and $P_{out}^{i}$, respectively. A new dataset is created using training and non-training class labels to train an attack model $\mathcal{Q}$, typically a binary classifier. A well-trained attack model $\mathcal{Q}$ can determine whether samples were involved in training the target model $M_{t}$.

The process of using MIA to determine the success of data unlearning is illustrated in Figure \ref{fig:mia}. After performing the unlearning operation on the client, the output of dataset $D_{f}^{i}$ on model $M_u$ is denoted as $P_{f}^{i}$. A judgment is made using the attack model $\mathcal{Q}$. If $\mathcal{Q}$ deduces that $D_{f}^{i}$ was not involved in training $M_u$, it indicates successful forgetting of $D_{f}^{i}$. Conversely, if $\mathcal{Q}$ determines that $D_{f}^{i}$ has not been forgotten, it implies that the unlearning process was not effective.

\subsection{Third-party data}
\label{sec:tpd}
The essence of our methodology lies in refining the model through the process of unlearning $D_{f}^{i}$ utilizing Equation \ref{eq:kla}, which, in turn, necessitates the incorporation of third-party data $D_{t}^{i}$. The guiding criterion for selecting $D_{t}^{i}$ is that it should not have been employed during the model's training. Two distinct approaches exist for the selection of this data:
\begin{enumerate}
    \item \textbf{Random noise} (Denoted as $\mathcal{D}_{rn}^{i}$): Consider $D_{t}^{i}$ as random noise, which is ensured to have not been involved in the model's training process. Obtaining random noise is a straightforward task, and during the unlearning process, it only needs to be dimensionally consistent with $D_{f}^{i}$.
    \item \textbf{Retain some client data from training} (denoted as $\mathcal{D}_{re}^{i}\subset D_{c}^{i}$): During the initial stages of training, clients reserve a portion of data without actively participating in the model's training process. When a client requires unlearning, this reserved data is used as $D_{t}^{i}$. In practical scenarios, clients may not necessarily use all available data for training the model, making it feasible to reserve a portion of data specifically for unlearning purposes.
\end{enumerate}

We conduct experiments using two distinct methods, selecting $D_{t}^{i}$ as $\mathcal{D}_{rn}^{i}$ and $\mathcal{D}_{re}^{i}$, to investigate the influence of third-party data selection on the model after unlearning. In Section \ref{sec:result}, we provide an in-depth discussion of the choice of $D_{t}^{i}$ and analyze the unlearning effects of these two methods based on the experimental results.
\begin{figure*}[!t]
    \centering
    \begin{minipage}[t]{0.23\textwidth}
        \begin{subfigure}{\textwidth}
            \includegraphics[width=\textwidth]{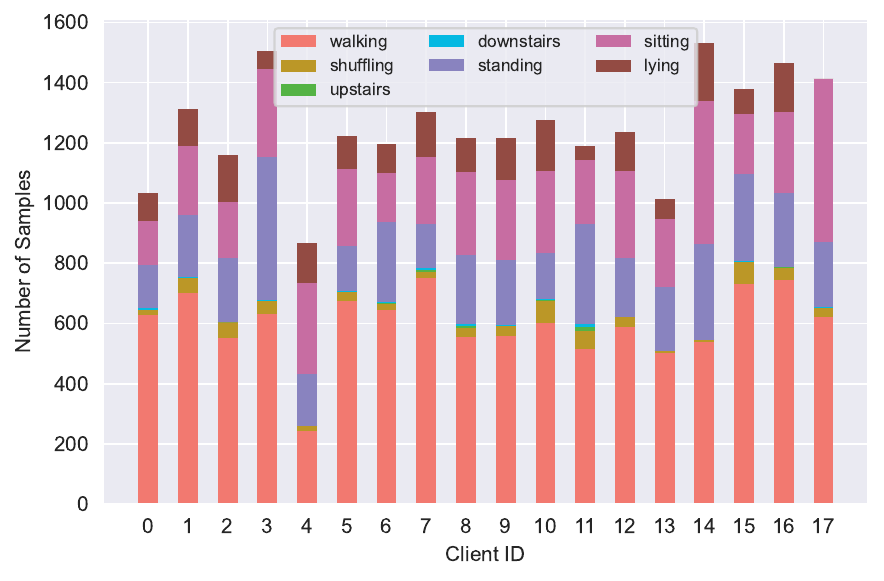}
            \caption{HAR70+}
        \end{subfigure}
    \end{minipage}
    \begin{minipage}[t]{0.23\textwidth}
        \begin{subfigure}{\textwidth}
            \includegraphics[width=\textwidth]{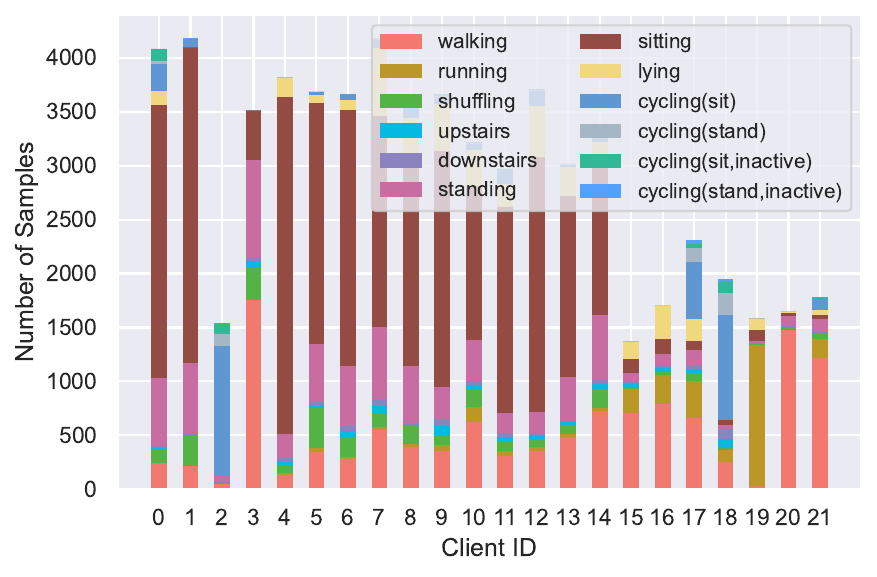}
            \caption{HARTH}
        \end{subfigure}
    \end{minipage}
    \begin{minipage}[t]{0.23\textwidth}
        \begin{subfigure}{\textwidth}
            \includegraphics[width=\textwidth]{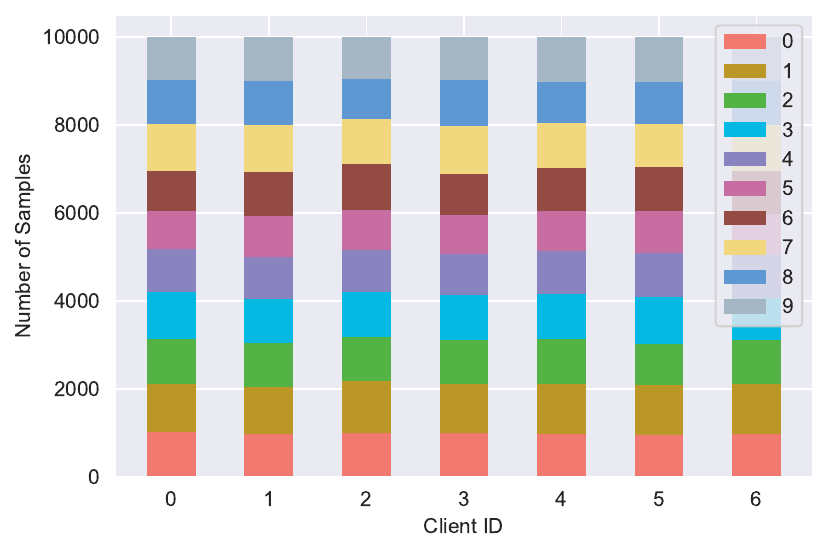}
            \caption{MNIST\textsubscript{iid}}
        \end{subfigure}
    \end{minipage}
    \begin{minipage}[t]{0.23\textwidth}
        \begin{subfigure}{\textwidth}
            \includegraphics[width=\textwidth]{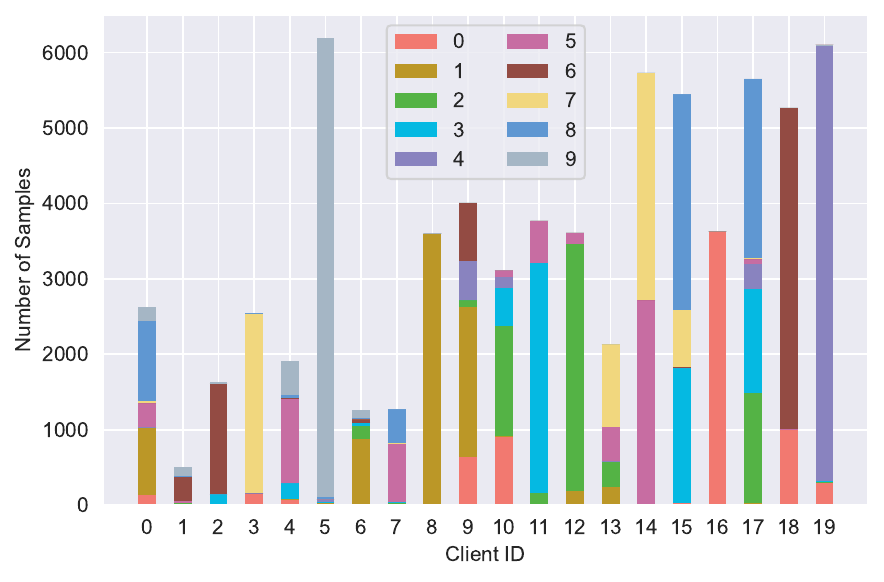}
            \caption{MNIST\textsubscript{non-iid}}
        \end{subfigure}
    \end{minipage}
    \caption{Data distribution for each client across various datasets, with distinct colors denoting different labels.}
    \label{fig:labeldistribution}
\end{figure*}

\section{Evaluation}
\label{sec:evaluation}
\subsection{Evaluation metrics}
\label{sec:metrics}
We use three metrics to evaluate the proposed methods, including \textit{MIA accuracy}, \textit{performance}, and \textit{time}.

\textbf{MIA accuracy}: Membership Inference Attack (MIA) is used to determine whether a sample has been included in the training of a model, thereby indicating its presence in the model's training set. MIA accuracy represents the percentage of correctly predicting whether a sample belongs to the model's training data. The MIA model is tested using an equal quantity of forgotten data $D_{f}^{i}$ and third-party data $D_{t}^{i}$ that were not used during training. Ideally, after unlearning, the MIA model should correctly identify $D_{f}^{i}$ as non-members of $D_{t}^{i}$, resulting in an ideal MIA accuracy close to $0.5$. MIA accuracy can be employed to assess the effectiveness of unlearning when a portion of the client data is forgotten.

\textbf{Performance}: The evaluation of an unlearning algorithm's performance involves assessing the model's performance after the unlearning process. Specifically, we examine the test accuracy of the model on the forgotten data $D_{f}^{i}$ and the remaining data $D_{r}^{i}$ from client $C_i$, as well as the test accuracy on each client's respective testing dataset. A reliable unlearning algorithm should successfully forget $D_{f}^{i}$ while preserving the model's performance.

\textbf{Time}: Time is a crucial indicator for measuring algorithm efficiency. One of the reasons for not utilizing the \textit{retrain} method for unlearning is its time-consuming nature. In the context of unlearning algorithms, the ability to achieve an unlearn effect similar to retraining in less time signifies a superior unlearning algorithm.

\begin{figure*}[!t]
        \centering
    \rotatebox{90}{\scriptsize{~~~~~~~~~~~~~~~reserved}}
    \begin{minipage}[t]{0.23\textwidth}
        \centering
        \includegraphics[width=\linewidth]{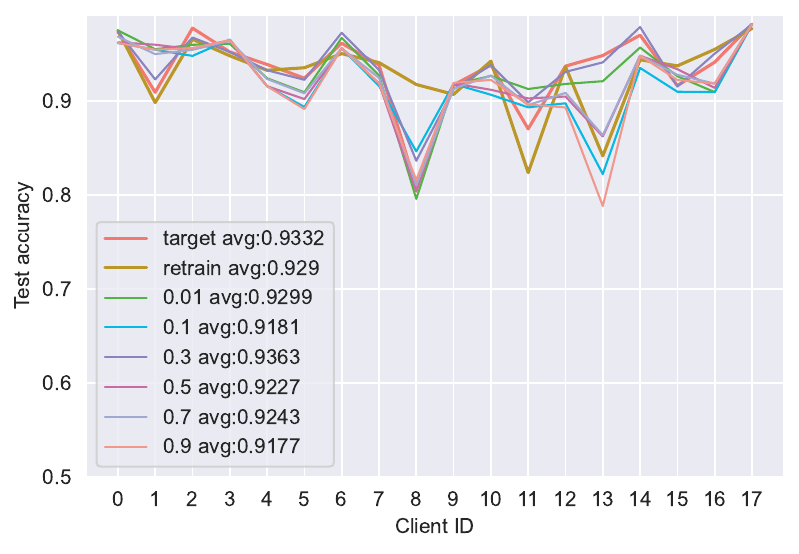}
    \end{minipage}%
    \begin{minipage}[t]{0.23\textwidth}
        \centering
        \includegraphics[width=\linewidth]{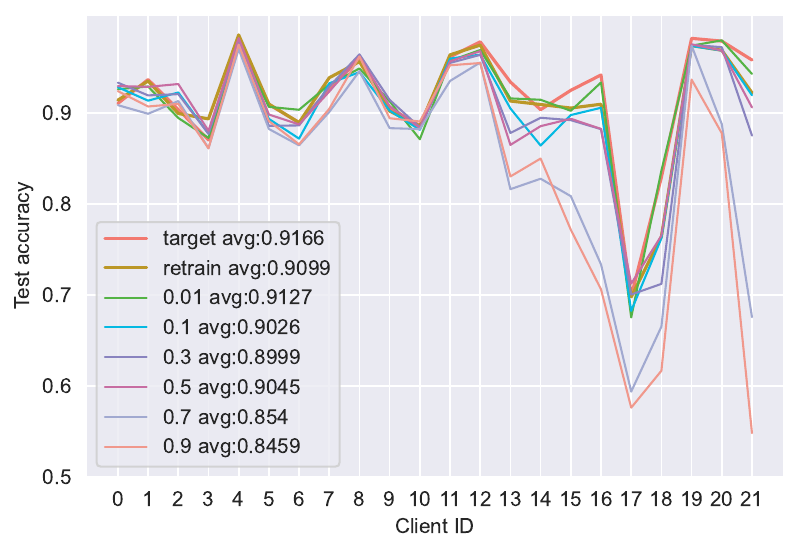}
    \end{minipage}%
    \begin{minipage}[t]{0.23\textwidth}
        \centering
        \includegraphics[width=\linewidth]{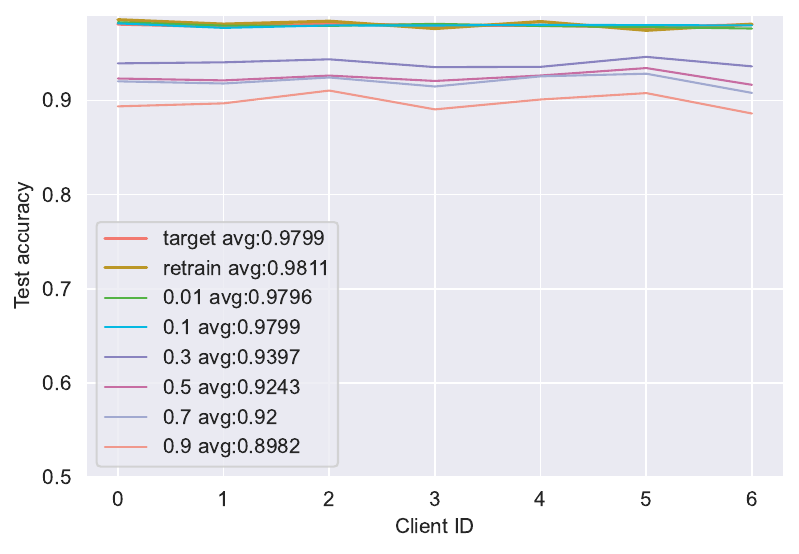}
    \end{minipage}%
    \begin{minipage}[t]{0.23\textwidth}
        \centering
        \includegraphics[width=\linewidth]{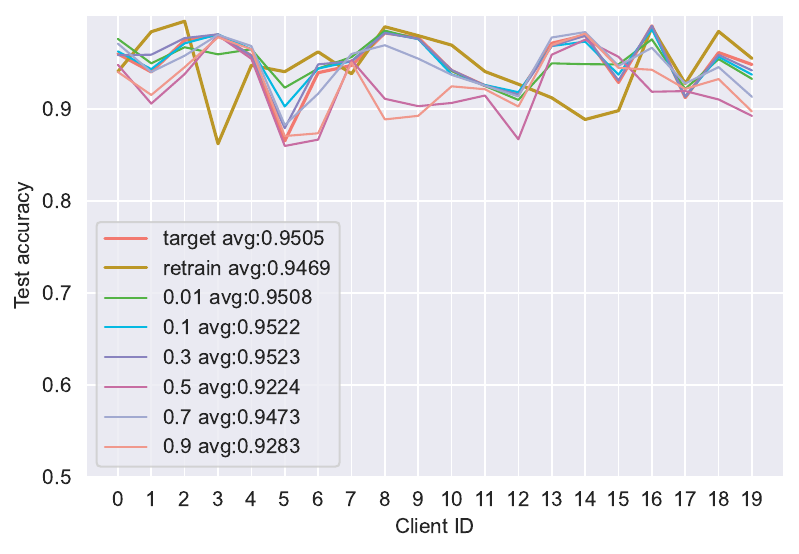}
    \end{minipage}

    \vspace{0cm} 
    \rotatebox{90}{\scriptsize{~~~~~~~~~~~~~~~~noise}}
    \begin{minipage}[t]{0.23\textwidth}
        \centering
        \includegraphics[width=\linewidth]{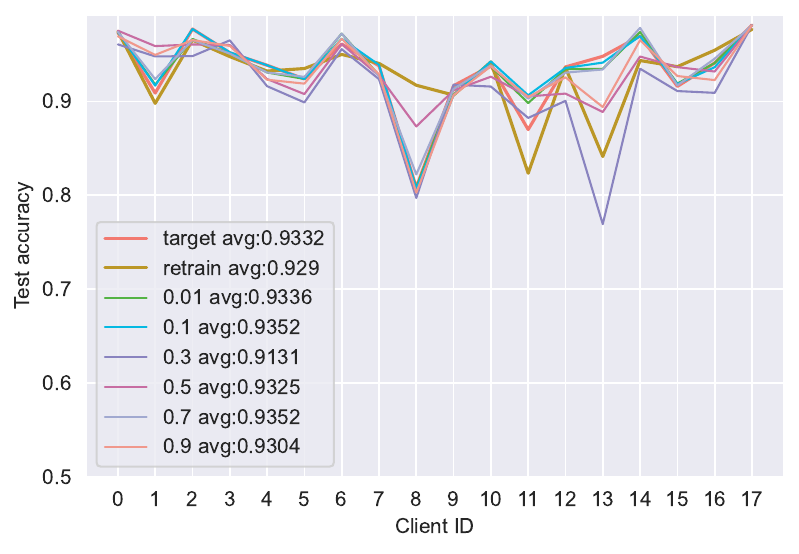}
        \subcaption{HAR70+}
        \label{fig:res-har70}
    \end{minipage}%
    \begin{minipage}[t]{0.23\textwidth}
        \centering
        \includegraphics[width=\linewidth]{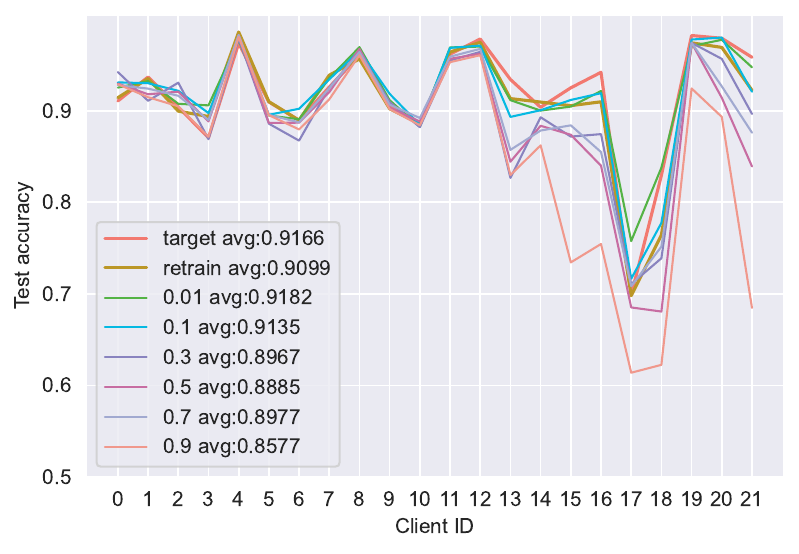}
        \subcaption{HARTH}
        \label{fig:res-harth}
    \end{minipage}%
    \begin{minipage}[t]{0.23\textwidth}
        \centering
        \includegraphics[width=\linewidth]{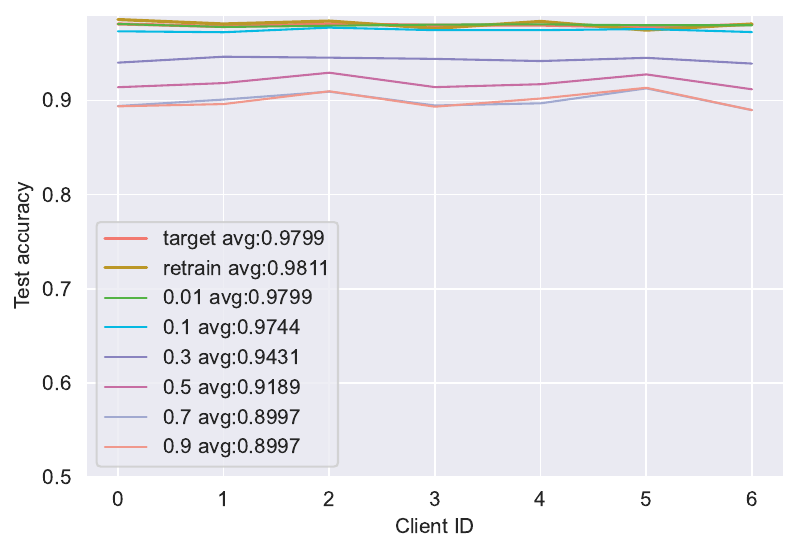}
        \subcaption{MNIST\textsubscript{iid}}
        \label{fig:res-iidmnist}
    \end{minipage}%
    \begin{minipage}[t]{0.23\textwidth}
        \centering
        \includegraphics[width=\linewidth]{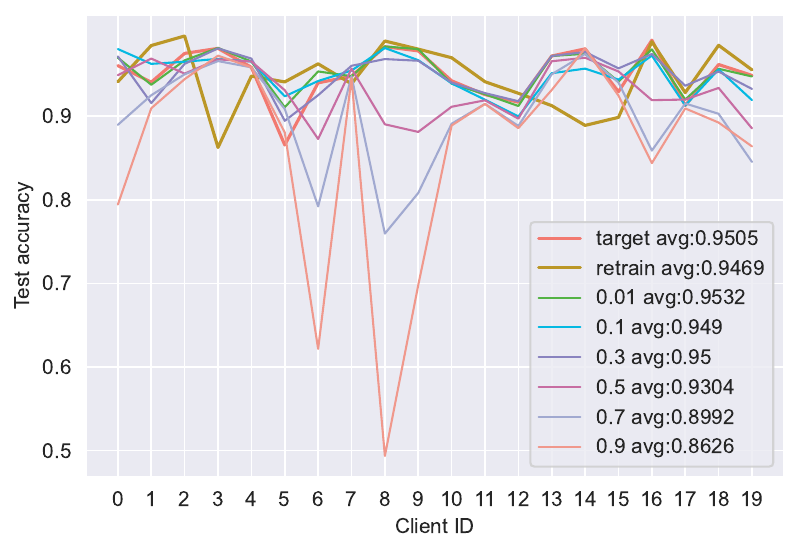}
        \subcaption{MNIST\textsubscript{non-iid}}
        \label{fig:res-non-iidmnist}
    \end{minipage}
    \caption{Performance evaluation of the unlearned model with different third-party data selection strategies.}
    \label{fig:result}
\end{figure*}

\subsection{Datasets}
\label{sec:datasets}
We conducted validation using two Human Activity Recognition (HAR) datasets: HAR70+ \cite{ustad2023validation} and HARTH \cite{logacjov2021harth}, \cite{bach2021machine}. In addition, to provide a comparative analysis with these HAR datasets, we also validated our approach on the widely used image dataset MNIST \cite{726791}. We included MNIST in our evaluation for several reasons. Firstly, we found that the test accuracy change of the model after unlearning in the HAR datasets was too insignificant to be persuasive. Therefore, we further validated our approach using the MNIST dataset. Secondly, the introduction of the MNIST dataset allowed us to apply both independent and identically distributed (iid) and non-iid processing. Lastly, we chose MNIST as one of the validation datasets to demonstrate the applicability of our method in other federated scenarios beyond the HAR domain.

The distribution of data samples among different clients for the four datasets is depicted in Figure \ref{fig:labeldistribution}. The MNIST\textsubscript{iid} dataset demonstrates clear iid characteristics, with consistent data distributions across each client. Conversely, the HARTH and MNIST\textsubscript{non-iid} datasets exhibit strong non-iid characteristics, as the data of each client significantly differs from one another. The distribution of data samples varies among individual clients. Although the HAR70+ dataset is not iid, the differences in data distribution between each client are not substantial.

\subsection{Model architecture and platform}

The experiments use the same model architecture for the two HAR datasets with two Conv layers and two FC layers, while the iid and non-iid MNIST datasets employ the LeNet-5 architecture \cite{726791}. All experiments are conducted on a single NVIDIA Tesla V100S GPU. Computation is performed on an Intel(R) Xeon(R) Gold 6240R CPU with 96 cores, running Ubuntu 20.04. The algorithms are implemented using PyTorch.

\subsection{Performance of the unlearned model}
\label{sec:result}

\begin{table}[!t]
    \begin{tabularx}{\linewidth}
        {X X >{\centering\arraybackslash}X >{\centering\arraybackslash}X >{\centering\arraybackslash}X}
        \hline
        \multirow{2}*{\textbf{Dataset}} & \multirow{2}*{\textbf{model type}} & \multicolumn{3}{c}{\textbf{Accuracy}} \\
        \cline{3-5}
        ~ & ~ & $D_{f}^{0}$ & $D_{r}^{0}$ & $D_{test}^{0}$ \\
        \hline
        \multirow{2}*{HAR70+} & target & 1.00 & 0.98 & 0.97 \\
        ~ & retrain & 0.97 & 0.99 & 0.97 \\
        \hline
        \multirow{2}*{HARTH} & target & 0.76 & 0.94 & 0.91 \\
        ~ & retrain & 0.45 & 0.96 & 0.91 \\
        \hline
        \multirow{2}*{MNIST\textsubscript{iid}} & target & 0.99 & 0.99 & 0.98 \\
        ~ & retrain & 1.00 & 0.99 & 0.99\\
        \hline
        \multirow{2}*{MNIST\textsubscript{non-iid}} & target & 1.00 & 0.94 & 0.96 \\
        ~ & retrain & 1.00 & 0.95 & 0.94 \\
        \hline
    \end{tabularx}
    \caption{Performance of the target and retrain models.}    
    \label{tab:performance1}
\end{table}

This section provides a detailed overview of the experimental results based on partial data from the unlearning client, denoted as $C_0$. The \textit{target} represents the results of unlearning the model, while the \textit{retrain} represents the baseline experiment of unlearning by retraining the model, as listed in Table~\ref{tab:performance1}. The \textit{unlearned} represents the experimental results of unlearning using our method, and $D_{test}^{i}$ denotes the test dataset of $C_{i}$.

\textbf{Accuracy on test datasets.} The results depicted in Figure \ref{fig:result} demonstrate the accuracy of the model after unlearning on each client's test dataset for different values of $\lambda$. It can be observed that in the HAR70+ dataset (Figure \ref{fig:res-har70}) and the MNIST\textsubscript{iid} dataset (Figure \ref{fig:res-iidmnist}), the deviation of each client's test dataset after unlearning, compared to the target and retrain results, is not significant. However, in the HARTH dataset (Figure \ref{fig:res-harth}) and the MNIST\textsubscript{non-iid} dataset (Figure \ref{fig:res-non-iidmnist}), which exhibit a stronger non-iid characteristic, some clients display significant deviations after unlearning, as evidenced by a notable drop in test accuracy for clients $C_{6}$ and $C_{8}$ in the MNIST\textsubscript{non-iid} dataset at $\lambda=0.9$.

\begin{table*}[!t]
    \begin{tabularx}{\linewidth}
        {X X >{\centering\arraybackslash}X >{\centering\arraybackslash}X >{\centering\arraybackslash}X X X}
        \hline
        \multirow{2}*{\textbf{Dataset}} & \multirow{2}*{\textbf{$D_{t}^{0}$ type}} & \multicolumn{2}{c}{\textbf{Accuracy}} & \multicolumn{2}{c}{\textbf{Time cost (second)}} & \multirow{2}*{\textbf{Speedup}}\\
        \cline{3-6}
        ~ & ~ & $D_{f}^{0}$ & $D_{r}^{0}$ & retrian & unlearn & ~ \\
        \hline
        \multirow{2}*{HAR70+}& reserved & 0.97 & 0.98 & 264.33 & 0.15 & {1762}$\times$\\
        ~& noise & 0.99 & 0.99 & 264.33 & 0.09 & {2937}$\times$ \\
        \hline
        \multirow{2}*{HARTH}& reserved & 0.59 & 0.95 & 1223.88 & 0.24 & {5099}$\times$ \\
        ~ & noise & 0.86 & 0.98 & 1223.88 & 0.20 & {6119}$\times$ \\
        \hline
        \multirow{2}*{MNIST\textsubscript{iid}}& reserved& 0.46 & 0.97 & 179.78 & 0.61 & {294}$\times$ \\
        ~ & noise & 0.22 & 0.96 & 179.78 & 0.18 & {998}$\times$\\
        \hline
        \multirow{2}*{MNIST\textsubscript{non-iid}} & reserved & 0.92 & 0.97 & 542.10 & 0.30 & {1807}$\times$ \\
        ~ & noise & 0.86 & 0.95 & 542.10 & 0.51 & {1062}$\times$ \\
        \hline
    \end{tabularx}
    \caption{Performance of the unlearned model with $\lambda=0.7$, where `noise' and `reserved' represent $D_{t}^{0}\in{\mathcal{D}_{rn}^{0}}$ and $D_{t}^{0}\in{\mathcal{D}_{re}^{0}}$, respectively.}
    \label{tab:performance2}
\end{table*}

The MNIST\textsubscript{iid} dataset demonstrates clear distinctions, as depicted in Figure \ref{fig:res-iidmnist}. With its iid characteristics, the data is uniformly and randomly partitioned among clients from a large dataset, resulting in a higher degree of similarity among clients and minimal differences. Following federated learning (FL), consistent accuracy is observed across clients when evaluated on the test dataset.

Despite the close similarity in data distribution among individual clients, the HAR dataset is unique in that data from each client is collected from different individuals. This inherent diversity among clients, even for the same activity, leads to variations between clients. Consequently, after FL, the model struggles to effectively generalize to each client, resulting in inconsistent performance across clients.

\subsection{Effectiveness of the unlearned method}
Considering unlearning effectiveness and model performance, we have selected $\lambda=0.7$ as an illustrative example for the experimental results in this section. By choosing $\lambda=0.7$, the algorithm prioritizes achieving better unlearning outcomes while also balancing the post-forgetting performance of the model to some extent.

\begin{figure}[!t]
    \centering
    \includegraphics[width=0.65\linewidth]{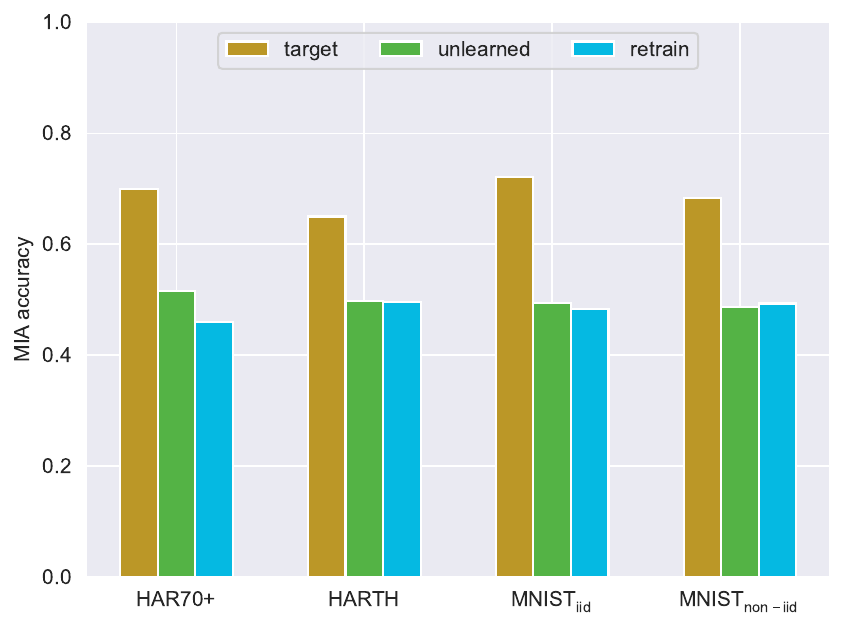}
    \caption{Membership inference result with $\lambda=0.7$.}
    \label{fig:res-mia}
\end{figure}

\textbf{MIA accuracy.} The assessment of unlearning success is based on Membership Inference Attack (MIA) accuracy. Figure \ref{fig:res-mia} illustrates the MIA accuracy on various datasets $C_0$ for unlearned data $D_f^0$ across the target model, the model after unlearning, and the model after retraining. It is evident that the MIA accuracy of the model after unlearning is lower than that of the target model, approaching the level observed in the retrained model. This observation suggests the successful removal of $D_f^0$ from the model through our proposed method.

\textbf{Performance and time cost.} In FL, the global model is formed by aggregating locally trained models from individual clients. However, the transmission of model-related parameters, such as weights and losses, between clients and the server incurs significant time and communication overhead. According to the data presented in Table \ref{tab:performance2}, the time required for retraining the model is orders of magnitude higher than that of fine-tuning methods. Our approach not only enables rapid unlearning of data but also mitigates the communication costs associated with model aggregation. Throughout the entire unlearning process, only three essential model parameter transmissions are required: 1) at the initiation stage, the server disseminates the latest global model $M_{t}$ to $C_{0}$; 2) after $C_{0}$ completes unlearning, it sends the forgotten model $M_{u}$ to the server; 3) upon receiving $M_{u}$, the server distributes $M_{u}$ to all clients.

\subsection{Impact of third-party data}
The results of unlearning $C_{0}$ with different third-party data selections are presented in Figure \ref{fig:result} and Table \ref{tab:performance2}. It is observed that, for both HAR datasets, the choice of different third-party data has minimal impact on the results. However, notable variations are evident in the case of the MNIST\textsubscript{iid} dataset. Specifically, in the MNIST\textsubscript{non-iid} scenario (refer to Figure \ref{fig:res-non-iidmnist}), as $\lambda$ increases, the test accuracy of $C_{6}$ and $C_{8}$ significantly drops when $D_{t}^{0}$ consists of random noise, compared to the scenario where $D_{t}^{0}$ retains client-specific data. Additionally, as shown in Table \ref{tab:performance2}, regardless of whether the MNIST dataset is iid or non-iid, when $D_{t}^{0}$ comprises random noise, the test accuracy on $D_{f}^{0}$ is notably lower compared to the scenario where $D_{t}^{0}$ retains client-specific data.

\section{Conclusion}
\label{sec:conclusion}
Our primary goal is to minimize system resource consumption while meeting the client's requirement for forgetting Human Activity Recognition (HAR) data. Here, we empirically showcase the effectiveness and feasibility of employing KL divergence for both model fine-tuning and unlearning client data. Additionally, we introduce a membership inference evaluation method to validate the efficacy of the unlearning process. Our approach is validated on two HAR datasets and the MNIST dataset with different data distributions. Across these datasets, experimental results demonstrate that our method achieves unlearning accuracy comparable to baseline methods, resulting in speedups ranging from hundreds to thousands. We aspire that our work will serve as a pioneering unlearning solution for HAR, contributing to enhanced privacy protection in this domain.


\end{document}